\documentclass[manuscript,screen,nonacm,review=false,timestamp=false]{acmart}
\usepackage{draftwatermark}
\usepackage{natbib}
\SetWatermarkLightness{ 0.9 }
\SetWatermarkText{Manuscript} 
\SetWatermarkScale{ 0.5 }
\AtBeginDocument{%
  \providecommand\BibTeX{{%
    \normalfont B\kern-0.5em{\scshape i\kern-0.25em b}\kern-0.8em\TeX}}}

\begin{document}
\title{Enhancing Instructional Quality: Leveraging Computer-Assisted Textual Analysis to Generate In-Depth Insights from Educational Artifacts}
\author{Zewei Tian}
\email{ztian27@uw.edu}
\affiliation{%
  \institution{University of Washington}
  \city{Seattle}
  \state{WA}
  \country{USA}
}

\author{Min Sun}
\affiliation{%
  \institution{University of Washington}
  \city{Seattle}
  \state{WA}
  \country{USA}}
\email{misun@uw.edu}

\author{Alex Liu}
\affiliation{%
  \institution{University of Washington}
  \city{Seattle}
  \state{WA}
  \country{USA}}
\email{alexliux@uw.edu}

\author{Shawon Sarkar}
\affiliation{%
  \institution{University of Washington}
  \city{Seattle}
  \state{WA}
  \country{USA}}
\email{ss288@uw.edu}

\author{Jing Liu}
\affiliation{%
  \institution{University of Maryland}
  \city{College Park}
  \state{MD}
  \country{USA}}
\email{jliu28@umd.edu}

\renewcommand{\shortauthors}{Tian et al.}
\begin{abstract}
  This paper explores the transformative potential of computer-assisted textual analysis in enhancing instructional quality through in-depth insights from educational artifacts. We integrate Richard Elmore's Instructional Core Framework to examine how artificial intelligence (AI) and machine learning (ML) methods, particularly natural language processing (NLP), can analyze educational content, teacher discourse, and student responses to foster instructional improvement. Through a comprehensive review and case studies within the Instructional Core Framework, we identify key areas where AI/ML integration offers significant advantages, including teacher coaching, student support, and content development. We unveil patterns that indicate AI/ML not only streamlines administrative tasks but also introduces novel pathways for personalized learning, providing actionable feedback for educators and contributing to a richer understanding of instructional dynamics. This paper emphasizes the importance of aligning AI/ML technologies with pedagogical goals to realize their full potential in educational settings, advocating for a balanced approach that considers ethical considerations, data quality, and the integration of human expertise.
\end{abstract}



\maketitle

\section{Introduction}
Instructional improvement is an iterative process where educators harness data-driven insights to refine the teaching practices and improve student learning. To meet the shifts in post-pandemic learning needs and the demand of artificial intelligence (AI) advancement on workforce development, the education system seeks new instructional and learning strategies that are personalized, effective, safe, and scalable \citep{cardona_artificial_2023}. Throughout the years, richer and more complex educational data have been generated by the advancement of instructional practices, providing vast potential for analyses but at the same time posing challenges to the approaches that process such data. Conventional quantitative methods are limited by the capacity of calculation and the efficiency of models, hence preventing efforts to improve teaching and learning outcomes. AI/ML approaches are able to effectively process the existing and forthcoming complex data with scalability and precision \cite{berland_educational_2014}, presenting an unprecedented opportunity to promote the research and instructional practices in education. These characteristics of new data and methods provide timely and actionable insights into the dynamics of the instructional environment. Furthermore, in recent years, this trend has been accelerated by the rapid adoption of generative AI tools, such as ChatGPT and Bard, which synergizes the capabilities of both text analysis and generation. A new field of research has emerged, in which researchers integrate the cutting-edge AI/ML techniques with educational domain knowledge of curriculum, teaching, and learning and to explore crucial questions for instructional improvement.

To ground the discussion of this paper into a clearly defined domain, we need to first provide general backgrounds and definitions of related terms. Artificial Intelligence, or AI, is about automation and generative capabilities. AI systems leverage hardware, algorithms, and data to generate ``intelligence," enabling tasks such as decision-making, pattern discovery, and the execution of various actions \cite{ruiz_glossary_2023}. Whereas on the other hand, Machine Learning (ML) is a range of approaches to develop algorithms that can identify rules and patterns inside structured or unstructured data including text, audio, video, etc. On the basis of ML, Natural Language Processing (NLP) specifically addresses the interpretation of linguistic data and helps computers manipulate such data to discover patterns \cite{chowdhury_natural_2003}. 

In this paper, we will review and discuss how AI/ML methods provide us with innovative solutions to analyze the textual data in education, as well as summarize the promises and pitfalls of these new methodological advancements. To guide our review of this emerging new field, we use the Instructional Core Framework that depicts the dynamic interactions among students, educators, and educational content to create learning opportunities and classroom environments \cite{city_instructional_2009, gillies_enhancing_2015, hennessy_analysis_2023}. Textual artifacts pertaining to each component and their interactions offer valuable insights into instructional practices. Specifically, we identify three main areas of adaptation of AI/ML analysis of textual data pertaining to three major components of instruction: teacher, student, and content. Each of these three components utilize and generate textual data at the same time. Teachers design curricula and lesson plans as textual data. At the same time, the instruction inside classrooms can also be transcribed to serve as data for analysis on instruction quality. On the other hand, teachers rely on the feedback from student assignments to provide tailored course content. Students submit their assignments as potential textual data for analysis while taking teachers' instruction as inputs. Content itself serves as a major textual data input, but also intakes new data like feedback from both of the other two components. For each component, we offer example studies and conduct a nuanced examination of the complexities, potentials, and limitations inherent in AI/ML-powered textual analysis of instructional artifacts. In the end, we remain cautiously optimistic about the value of this line of research in advancing education and identify several major directions for future research.

\section{Instructional Core Framework}
We use the Instructional Core framework to facilitate the categorization of textual analysis in educational research, analyze the use cases, and anticipate future directions.  We also use this framework to contextualize the findings and assist the audience to understand the broader implications of the results within the existing body of knowledge. The instructional Core framework  challenges prevailing paradigms, advocating for a departure from teacher-centric instructional approaches towards a more dynamic interplay of pivotal constituents \cite{city_instructional_2009}. Teachers, students, and content converge as the central components in these intricate pedagogical interactions. The instructional core functions as a guide through the multifaceted terrain of instruction, offering an analytic lens into the classroom dynamics that are conducive to the enhancement of both teaching and learning.

\begin{figure}[ht]
  \centering
  \includegraphics[width=\linewidth]{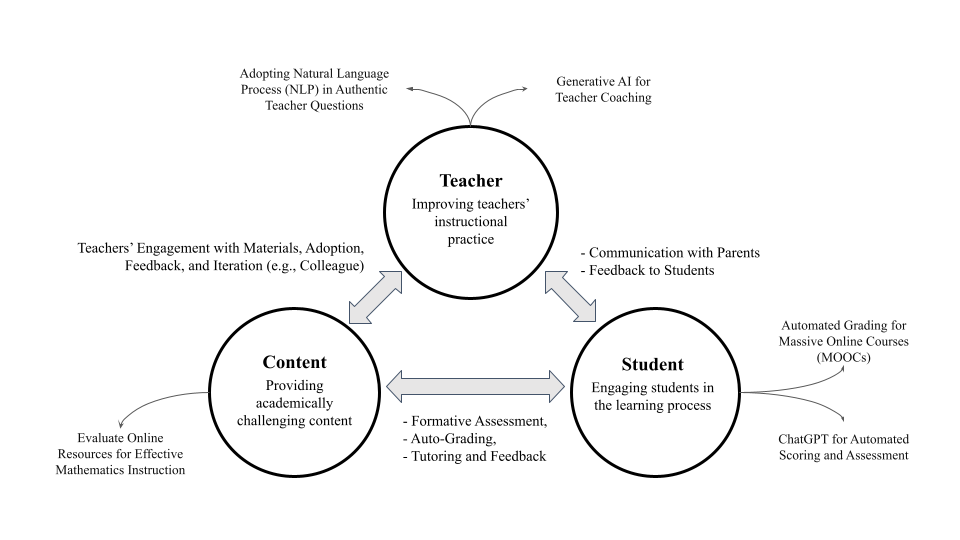}
  \caption{Instructional Core Framework by~\citet{elmore_improving_2008}, with examples of educational artifacts.}
  \label{ICF}
\end{figure}

The instructional core includes three interdependent components, with dynamic interactions between one another. These components include \emph{teachers}' knowledge and skill, \emph{students}' engagement in learning, and academically challenging \emph{content}. The components in the instructional core are decisive of educational performances and achievements as the desired outcomes that improve the quality of teaching and learning. To change performance, either the components themselves or the relationship between teacher and the student in the presence of content need to be changed. It is indicated that basically only three ways of increasing learning and performance exist: increasing the knowledge and skill of teachers, affecting content, and altering the relationship between the components \cite{elmore_improving_2008}. These approaches aim to make the instructional cores more practical and implement them effectively.

Instructional improvement can focus on any of the three components in the instructional core, including the quality and dynamics of the interactions between them. For example, efforts for improvement in these three components of instructional core can involve providing feedback on teachers' knowledge and practices \cite{bain_technology_2011}, academic assessment based on group and individual student practice opportunities \cite{doabler_components_2019}, and generating high-quality and coherent curriculum and materials as instructional content \cite{harris_impact_2015}. These efforts aim to enhance the effectiveness of teaching and learning processes, contributing to instructional improvement in educational settings. 

Besides the individual components, the connections between instructional core components are also key characteristics integrated within the framework, linking teachers, students, and content in a holistic manner. These interdependent relationships provide additional ways to improve educational practices as well, leading to another important aspect in the instructional core framework which is that once you change one component, you have to change them all \cite{city_instructional_2009}. As indicated in the paradigm, all three components are correlated, meaning that disentangling one from the other two may cause imbalance and consequently fail to create desired outcomes of improving instructional practices. If we aim to enhance the content, we should assess whether teachers are equipped to handle the more advanced curriculum and also consider whether students can effectively engage with it \cite{elmore_leading_2010}. This demonstrates the interplay between content and teachers, as well as between content and students. Merely incorporating high-level content into the curriculum without taking into account the acceptance of both students and teachers would not be prudent. For instance, to foster student engagement, it is crucial to furnish students with a suitable curriculum while ensuring that teachers are capable of managing varying levels of student achievement and maintaining classroom control. The perspectives may differ, but they all contribute to a comprehensive evaluation of all three components and their interrelationships.

To achieve effective instructional improvement, it is crucial to thoroughly gather, organize, and analyze relevant data pertaining to the instructional core. As mentioned, the volume of educational data generated by innovative instructional activities in recent years poses new challenges. With the help of AI/ML, this integrated process enables educators and administrators to gain valuable insights and make informed decisions based on evidence. There are different forms of textual artifacts of classroom instruction generated, such as transcripts of instructional discourses, students' written homework, chat history between tutors and students, textbooks and different types of Open Educational Resources. These first-hand educational materials and data contain nuanced details and patterns of classroom dynamics which are often absent from traditional, structured, administrative data produced by conventional methods \cite{abrahamson_learning_2016, scribner_dynamics_2001}. NLP and other methods can analyze them timely. Then, the integration of education domain knowledge to guide these methods applications will be the key. We envision this type of method will gain momentum and importance moving forward. The rest of this paper will demonstrate how such integration would benefit educational practices from the perspectives of three instructional cores, introducing the AI/ML methods adopted and how the combination with education domain knowledge is implemented. 

To show how the integration of AI/ML in the instructional core holds immense transformative potential for education in the future, revolutionizing teaching practices and enhancing student learning experiences \cite{cardona_artificial_2023}, here we provide an overview of the case studies and beyond. By providing automated feedback on teacher discourse, AI/ML becomes a powerful tool for professional development and instructional change. For students, it facilitates autograding and coursework evaluation, offering valuable feedback and access to intelligent tutoring systems. Additionally, AI/ML enables large-scale data analysis, providing unique insights and assisting in the generation of tasks, lesson plans, and other course materials. This integration not only streamlines administrative tasks but also reshapes the dynamics within the educational landscape. In terms of teacher-content dynamics, AI/ML offers tailored tasks suited to varying cognitive demands while providing teachers with manageable plans. It bridges the gap between teachers and students by diminishing pre-existing perceptions of student achievement levels. This transformation eliminates the divide between teachers' assumptions, research-based expectations, and students' actual capabilities, fostering effective conversations and facilitating a more accurate understanding of students' potentials. Ultimately, the integration of AI/ML in the instructional core has the power to propel education towards unprecedented possibilities, fostering deeper connections and personalized learning experiences.

\section{Textual Analysis Aims to Support Teachers' Learning}
In recent years, the landscape of professional development for educators has undergone a transformative shift, leveraging the capabilities of AI and machine learning to enhance instructional practices. Traditional feedback mechanisms, often sporadic and evaluative in nature, are evolving as technology becomes an integral part of this process \cite{demszky_can_2023, jensen_toward_2020}. As part of practical demand, research has just started to demonstrate the efficacy of AI-driven tools in refining teacher discourse and fostering a more inclusive classroom environment. Under this context, NLP offers a powerful approach to measure teaching practices through the analysis of textual data like classroom transcripts by identifying linguistic nuances, communication styles, the frequency and quality of teacher-student engagement \cite{tosey_neurolinguistic_2010}. This provides educators with valuable insights into their strengths and areas for improvement. In this section, we summarize a few exemplary studies. 

\subsection{Advancing Classroom Discourse: Adopting Natural Language Process (NLP) in Authentic Teacher Questions}
Classroom discourse in educational settings manifests through diverse forms, ranging from teacher-directed lectures and procedural communication to open-ended discussions fostering collaborative exchanges between students and teachers \cite{alexander_culture_2008, juzwik_inspiring_2015}. By creating dialogic space and opening up classroom discourse, teachers are able to enhance students' active learning, particularly in language classrooms, as high-quality classroom discourse is characterized by open and authentic questions along with formative feedback whereby student contributions are probed and elaborated on \cite{hardman_opening-up_2016, soter_what_2008}. 

The abundance of authentic questions can promote substantive conversation and are related to student engagement and achievement growth in a broader sense. \citet{kelly_automatically_2018} incorporated AI to address the insufficiency of scaling the measure of question authenticity, which previously relied on human observations or coding. The study adopted automatic speech recognition, NLP, and ML to train models to detect authentic questions. The methodology for this study underwent iterative refinement, drawing insights from two primary data sources. Firstly, a comprehensive archival database with text transcripts from 451 observations across 112 classrooms was utilized (Partnership data). Additionally, a new set of 132 high-quality audio recordings from 27 classrooms was collected (Class 5.0 data), adhering to technical constraints aimed at anticipating large-scale automated data collection and analysis. These diverse data sources laid the foundation for the investigation conducted by the research team. 

The research adopted different methods for different data collections \cite{kelly_automatically_2018}. The first one is a semi-automated approach on the Partnership data, relying on human coders to deliver a wide range of different tasks including segmenting the raw classroom audio into teacher utterances, identifying which utterances are questions, distinguishing instructional questions from non-instructional questions, and providing approximate transcriptions of the instructional questions. The next step includes syntactic and discourse parsing \cite{manning_stanford_2014, surdeanu_two_2015} which was used to compute sentence (a teacher utterance) and multi-sentence discourse features. The study team measured various properties of language to arrive at a set of 244 features at each level of structure (word, sentence, discourse), including question stems (e.g., ``what"), word order, and referential chains, etc. The second one, applied on the Class 5.0 data, was a fully automated approach with the ultimate goal of providing a measure of authenticity from a recording of classroom audio without any human involvement. 

With different approaches and abundant datasets collected, a sufficiently high correlation between the computer- and human-coded authenticity suggests a promising avenue towards more automated measurement. It highlights how NLP techniques can be used for such analysis purposes, not only offering a more efficient and scalable approach to measuring question authenticity but also holding the potential for improving teacher professional development and classroom instruction. However, the authors also recognized the challenges that may occur when implementing in real-world cases, which include issues like noise, dialect diversity, data imbalance, and many more. It is undeniable that authenticity in itself is a complex construct, yet still, the inclusion of automated approaches empowered by NLP can offer some promising solutions to identify authentic questions. By leveraging NLP, innovative approaches can provide high-quality feedback to teachers derived from transcripts of classroom interactions and detect instructional factors which are well-aligned with commonly used observation protocols \cite{danielson_evaluation_2013}. This unlocks the potential to enhance existing classroom observation systems through collecting far more data on teaching with a lower cost, higher speed, and the detection of multifaceted classroom practices \cite{liu_measuring_2021}.

Accountable talk moves are effective probes in mathematics classes with amplified student achievement \cite{michaels_deliberative_2008}. By utilizing advances in automatic speech recognition and NLP, automated approaches can help identify such patterns to direct classroom discourse into a more engaging and efficient way \cite{jacobs_promoting_2022}. The pilot study showcases a promising trend of increased talk moves, emphasizing the potential of platforms like the TalkMoves application to democratize access to high-quality feedback for educators. Building on these advancements, the prowess of ML in swiftly and accurately classifying classroom discourse activities is disclosed, offering automated insights that align remarkably well with human-coded evaluations \cite{wang_automatic_2014}. This convergence underscores the capacity of NLP to streamline feedback processes and provide invaluable support for instructional improvement by utilizing textual data generated through the instructional practices. 

The instructional processes in classrooms consist of a diverse set of activities, one of which is ``lesson study" where teachers systematically examine their practice and improve their instructional practices \cite{makinae_origin_2019, saito_key_2012}. ``Lesson study" recognizes the complexity of cognitive demands inside classrooms and provides ways for teachers to understand such complexity. With the assistance of AI/ML, the activities can be operated in a consistent and fluent manner. Appropriated trained and fine-tuned NLP models would efficiently analyze huge amounts of data generated from inside the classrooms and provide accurate feedback for teachers. With guidance from these analyses and feedback teachers can now develop their own learning and understanding, helping both teachers and students achieve high-level tasks in classrooms which consequently predicts higher performance and cognitive skills. By having the high-level tasks and the right instructional practice, students are capable of achieving very competitive outcomes \cite{newmann_authentic_2001}.

\subsection{Generative AI for Teacher Coaching}
Teachers' actions and inactions in the classroom become critical in determining the quality of the learning experiences for students and their well-being. Professional development serves as a dynamic and evolving mechanism through which educators engage in continuous learning and refinement of their skills. Internationally, classroom observation is widely recognized as a key tool for teachers' professional development and evaluation \cite{martinez_classroom_2016}. In the loop of such evaluation, the feedback is conventionally provided by school administrators, peer teachers, and instructional coaches. There are also pre-determined protocols and assessments that include various rubrics, facilitating the evaluation process. However, due to the variation in expertise, resources, and human subjectivity under different contexts, the assessments of such kinds inherently have limitations in their validity and applicability \cite{kelly_using_2020}. The heterogeneity of evaluation processes and feedback indicates disparities in teacher development opportunities and practices, which in turn significantly impact student educational outcomes.

Prior efforts have shown the promising characteristics of adopting machine learning and natural language processing in assessing teachers' instructional effectiveness such as detecting authentic questions \cite{kelly_automatically_2018} and accountable talk moves \cite{jacobs_promoting_2022, suresh_talkmoves_2022}. These works mostly include detection and recognition of key elements that have positive impacts on student learning, thus selected to annotate the areas for improvement. Such tasks can be easier to get automated to provide fast and accurate feedback on instructional practices, but are still subject to human interpretation for substantial works of improvement. To provide high-level, actionable feedback similar to the ones proposed by human evaluators and coaches during professional development sessions, generative AI, like GPT-3.5, has huge potential in providing relevant feedback that aligns with conventional coaching. These tools have already demonstrated capabilities in generating texts in education settings, and are widely adopted by teachers and school professionals, playing multiple roles in pedagogical activities \cite{jeon_large_2023}. 

In the selected case study for this section, researchers investigated whether ChatGPT can help teachers and education professionals by providing effective feedback on a high level, like generating classroom observation rubrics and helpful pedagogical suggestions \cite{wang_is_2023}. The study was conducted three-fold, assigning three separate tasks to ChatGPT. The first one is to score a transcript segment for items derived from classroom observation instruments, according to criteria of the Classroom Assessment Scoring System (CLASS) and Mathematical Quality Instruction (MQI) rubrics. The second one is to identify highlights and missed opportunities for CLASS and MQI items, elaborating on these elements that occurred in the process of instructional practices. The third one is to provide open-ended, actionable suggestions to the teacher for eliciting more student mathematical reasoning in the classroom. These three stages of task analysis are similar to what evaluation and coaching would look like with human to human interactions, but are on a more efficient basis because of the automated process. 

The data used in this study was extracted from the National Center for Teacher Effectiveness (NCTE) Transcript dataset \cite{demszky_ncte_2023} in this work, the largest publicly available dataset of U.S. classroom transcripts linked with classroom observation scores. The dataset consists of four years (2010-2013) of classroom transcripts from 4th and 5th grade elementary mathematics observations. At the time of data collection, experts assessed and rated the transcripts based on two instruments, both CLASS and MQI. CLASS instrument segmented instructions into 15-minute sections, whereas MQI corresponds with 7.5-minute sections. All randomly selected segments are provided to ChatGPT in combination with zero-shot prompting. The authors measured zero-shot performance firstly because segments of the transcripts are long, causing annotated sections to exceed the input limit. Additionally, zero-shot prompting closely aligns with teachers' day-to-day interactions with ChatGPT, utilizing the generative AI as it is given without specific fine-tuning. The study recruited human participants to examine the validity of ChatGPT's performance on all three tasks, evaluating the outputs alongside human ratings. The results indicated that ChatGPT is able to provide relevant responses, yet failed to give novel and insightful feedback. This can be attributed to the scarcity of data, leading to insufficient information around examples of teacher coaching. The incorporation of generative AI presents promising prospects for attaining entirely automated teacher coaching instructions and feedback, contingent upon the availability of ample training data and illustrative examples. While it is acknowledged that zero-shot performances fall short of constituting a flawless solution, it demonstrates both the potential and the areas for improvement.

In sum, this study employs a zero-shot performance evaluation approach, mimicking how teachers might interact with generative AI in their day-to-day activities without specific fine-tuning. This approach aligns with practical implementations and offers insights into the model's adaptability. However, the effectiveness of generative AI, as demonstrated in the study, is contingent upon the availability of ample training data and illustrative examples, the lack of which points out challenges in replicating the novelty and insightfulness of human-generated feedback. Despite  illustrating ChatGPT's performance of zero-shot prompting, the study still calls for development and fine-tuning required for AI models in the context of teacher coaching and professional development.

\section{Textual Analysis for Student Support and Assessment}
AI/ML technologies including NLP present diverse avenues for enhancing students' coursework. This transformative potential is evident in three major applications: (a) generating personalized and adaptive items through fine-tuned models tailored to specific subject areas; (b) analyzing and auto-grading coursework; and (c) providing students with feedback, including intelligent tutoring. Just as AI/ML contributes to the democratization of high-quality feedback for educators in professional development, its applications in supporting student learning  hold the promise of more personalized, adaptive, and efficient learning experiences for students. 

Integrating AI/ML in adaptive learning emerges as an approach to customize instruction to learners' background, experiences, and prior knowledge \cite{alam_harnessing_2023, peng_personalized_2019, walkington_using_2013}. It possesses the capability to recommend optimal content from an ample amount of available resources, provide guidance on well-structured long-term curricula, and facilitate the connection of suitable learners through suggestions and precise performance evaluations, among other functionalities \cite{maghsudi_personalized_2021}. The range of customization spans across different subject areas, as well as various means of pedagogical designs \cite{ruiz-rojas_empowering_2023}. 

Furthermore, teachers have used AI/ML-powered tools to grade assignments, provide efficient and timely feedback to students, and predict student learning trajectory and outcome \cite{chen_predicting_2018, wilson_classification_2022}. This automation of assessment processes offers benefits such as rapid evaluation, consistency in grading, and the ability to handle large volumes of student work. This type of technology  takes input from student submissions through online platforms or through transcribed responses to conduct computer-based algorithms like text mining in order to find the similarities between student responses and the assessment requirements determined by teachers in grading criteria \cite{kakkonen_automatic_2004}. By targeting the implementation of AI/ML methods to free-text questions for testing the comprehensive ability of students, teachers can be alleviated from heavy workload with the assistance of automated grading systems which focus on the critical words and sentences, analyze the logical semantic relationship of the context and predict the interpretable grades \cite{wang_intelligent_2018}. Even without the grading rubrics and building NLP models by only taking human graded answers into consideration, some currently available auto-grading technology can still handle the assessment and achieve a good inter-rater agreement with expert grading \cite{andre_can_2017}. 

The third application, intelligent tutoring, is a dynamic combination of the previous two. By leveraging student log data and utilizing sentence‐level semantic representations of student responses to open‐ended questions, AI/ML can provide a collaborative filtering‐based approach to both predict student scores as well as recommend appropriate feedback messages for teachers to send to their students \cite{botelho_leveraging_2023}.  Intelligent Tutoring Systems (ITSs) have been central to deliver adaptive guidance and instruction, evaluate learners, define and update the learner's model, and classify or cluster learners \cite{corbett_chapter_1997, mousavinasab_intelligent_2021, nwana_intelligent_1990}.

\subsection{Automated Grading for Massive Online Courses (MOOCs)}
The migration from physical materials to digital resources has been predominant over the past years, especially after the impact of the pandemic. The fact that more and more students are relying on computers to complete and submit their school work requires novel and innovative approaches for teachers to adapt to new media when it comes to grade assignments and assess student performances. One study provides a comprehensive exploration into the domain of automated grading tools, particularly focused on essay writing and open-ended assignments within the context of Massive Online Courses (MOOCs) \cite{zhao_memory-augmented_2017}. The research addresses the burgeoning need for scalable grading solutions as more students rely on digital platforms for academic submissions. The proposed model, centered around memory networks, presents a novel approach wherein graded samples for each score in the rubric are utilized to predict scores for similar, ungraded responses. Notably, the model demonstrates state-of-the-art performance, specifically excelling in 7 out of 8 essay sets, as evidenced by evaluation on the Kaggle Automated Student Assessment Prize (Kaggle ASAP) dataset. 

The authors exclusively employed the Kaggle ASAP dataset, leveraging it for both training and evaluating the memory networks-powered automated grading model. For the model of the study, it comprises four key layers: Input Representation, responsible for generating vector representations of student responses; Memory Addressing, loading selected responses into memory with weighted assignments; Memory Reading, retrieving content based on weighted summation; and Output, making predictions from the resulting state. Stacking Memory Addressing and Reading layers enhances the model's ability to learn abstract representations through successive computational stages. The primary focus lies in establishing a robust and reliable representation of assignments through vectorization, coupled with the storage of pertinent samples in the memory component. However, the study recognizes the need for future endeavors to expand the model's applicability, urging the exploration of diverse datasets featuring varied assignment formats to enhance generalizability. Attention is also directed toward refining assignment representation and mechanisms for measuring relevance among assignments in forthcoming research. 

By comparing the model with Enhanced AI Scoring Engine (EASE), an open-source AES system, to demonstrate the improvements on performance, the study emphasizes the commendable success of the memory-augmented neural model in achieving high performance, particularly for essay sets. The authors indicate that there are two key factors to the performance: reliable representation and memory component. By recognizing and accounting for these two factors, the model's potential generalizability to assignments from diverse subjects underscores its significance in the broader educational landscape. Acknowledging its merits, the study concurrently underscores its limitations, notably being tested solely on the Kaggle ASAP dataset. This calls for further exploration with diverse datasets containing various assignment formats. The other area marked for improvement is the representation of the assignment and the mechanism for measuring relevance among assignments, requiring enhancement on these aspects to improve the overall robustness and applicability of the model.

\subsection{ChatGPT for Automated Scoring and Assessment}
Generative AI like ChatGPT utilizes LLMs to generate responses in accordance with prompts and inputs. By further fine-tuning the models, especially when such approaches are provided and integrated into the service, LLMs can demonstrate higher levels of performance specifically designed to tackle different tasks. For example, with sufficient training data, a 2-billion parameter model named MathGLM can provide multi-digit arithmetic operations with almost 100\% accuracy without data leakage, significantly surpassing GPT-4 \cite{yang_gpt_2023}. Beyond this, generative AI are adopted under other contexts including automatically scoring student-written constructed responses using example assessment tasks in science education. 

Others investigate the application of fine-tuned ChatGPT (GPT-3.5) for this specific task \cite{latif_fine-tuning_2023}. While GPT-3.5 has demonstrated proficiency in natural language processing, its direct use for scoring student responses is limited by the contextual variation in language. The study engaged in a fine-tuning process, training GPT-3.5 on specific assessment tasks using a dataset that included student responses and expert scores. The research encompassed six tasks, each characterized differently, to comprehensively assess the model's performance. A comparative analysis was conducted, comparing GPT-3.5 with BERT— which stands for Bidirectional Encoder Representations from Transformers, another pre-trained natural language processing model that has significantly advanced language understanding tasks by capturing bidirectional context representations \cite{devlin_bert_2019}. The research aimed to gauge the effectiveness of GPT-3.5 in enhancing automatic scoring accuracy. Additionally, the study explored text data augmentation as a creative strategy to leverage GPT-3.5-turbo's capabilities for enhancing automated evaluation precision. It is demonstrated that the effectiveness of this approach in producing a more extensive and diverse training set for scoring mechanisms, leading to improved performance in evaluating student text responses \cite{cochran_improving_2023}. Furthermore, the study adopted a benchmark PandaLM— which is a tool designed for optimizing domain-specific training and refining large language models' instructions \cite{wang_pandalm_2023}— to facilitate a comprehensive evaluation of model performance across various tasks. The results indicate that fine-tuned GPT-3.5 outperformed BERT in scoring accuracy across all six tasks. GPT-3.5 demonstrated a remarkable average increase of 9.1\% in accuracy for the multi-label and multi-class tasks, affirming its effectiveness in domain-specific automatic scoring. The findings suggest that domain-specific fine-tuning enhances the performance of language models for educational assessment tasks, providing a valuable tool for educators and researchers. The fine-tuned models have been released for public use and community engagement, which further supports its practical implications.

Despite great promise, potential limitations may include the need for extensive fine-tuning and a broader generalizability study of model performance and utility across various educational contexts, like online/in-person instruction or class sizes, and subjects.

\section{Textual Analysis for Content Analysis and Development}
Data science technologies and artificial intelligence can provide unprecedented opportunities for educational research. The capabilities of analyzing and generating text data can be incorporated into the content aspect of Elmore's Instructional Core Framework. Specifically, natural language processing can be applied to massive amounts of text data and by analyzing them, can provide distinct insights which previously were neglected. Additionally, generative AI enables teachers to pay more attention to working with students, as these specifically trained language models can relieve them from many routine tasks like lesson planning.

Cultivating the power of AI/ML with text data is a promising approach to analyze student essays, forum discussions, and other educational text data to gain insights into learning processes, identify common misconceptions, and improve learning outcomes. Textbooks are widely utilized as learning and instructional technologies \cite{torney-purta_citizenship_2001}, and are swiftly evolving along with the trends of technological development. Furthermore, high-quality textbooks and instructional materials in general can have significantly positive effects on student learning and outcomes \cite{read_where_2015}. Large language models (LLMs)—a type of language model that can both understand and generate texts—is an integrated approach to both analyze the previous resources and provide new materials based on insight gained from the analyses. Drawing content from 15 U.S. history textbooks, researchers examined the language using several methods to identify the prevalent topics of textbooks, the actors discussed (gender, ethnicity) and their characterization (as passive or active via lexicons), and the contexts they are related to \cite{lucy_content_2020}. Besides running analyses on previous textbooks and educational resources, LLMs and generative AI show promise in other applications, including in-time evaluation on online educational resources, which are accumulating swiftly and providing new insights.

\subsection{Leveraging Technology and Expertise to Evaluate Online Resources for Effective Mathematics Instruction}
As the online landscape of educational resources burgeons, schools and educators are increasingly turning to these materials alongside traditional textbooks. Open educational resources (OERs), with their undeniable promise of cost-effectiveness and accessibility, present a prospect for overcoming the educational gap and fostering educational justice by democratizing the access to high-quality materials \cite{richter_open_2012}. However, a crucial question looms: how can we ensure the pedagogical efficacy and curricular alignment of these readily available resources, particularly in the critical developmental stage of mathematical learning at a given age? 

An interdisciplinary group of researchers, funded by National Science Foundation, proposes a novel and rigorous approach to addressing this critical gap by integrating domain knowledge in mathematics instruction with the latest technologies of AI/ML analysis of text data. 

\subsubsection{Open Education Resources as Data}
The study begins with an extensive data collection phase, amassing over 300,000 math lesson plans from a variety of OER platforms, including widely known sites like \emph{BetterLesson}, \emph{IllustrativeMathematics}, and \emph{Achieve the Core}. The research team pays special attention to the representativeness of the sampled lesson plans to guarantee that the collection exemplifies the breadth of resources available for educators. The gathered lesson materials are methodically organized and stored using advanced data management solutions, enhancing efficiency in data handling and allowing for the integration of various types of data, such as download numbers and comments from the OER websites. The team also conducts regular exploratory data analysis to examine the distribution of key measures within the lesson materials. This process is crucial for making sense of the data and identifying any potential errors. This robust data collection and analysis formed a solid foundation for the subsequent phases of the study, where the quality of these lesson materials are scrutinized and enhanced through human-centered evaluations and AL/ML algorithms. 

\subsubsection{Classification methods of high quality lesson plans: human-centered data science approach}
Central to the study is a human-centered method for assessing the quality of lesson plans, an approach instrumental in ensuring the reliability and validity of both the measures and the algorithms employed. This approach initiates with the development of a theory/conceptual framework for high-quality instruction through comprehensive literature review and the application of the Delphi method, a mixed-method technique combining surveys and an expert panel to achieve consensus on key aspects of instructional quality and their measurement. Utilizing domain expertise from educators and researchers, the study formulates rigorous quality measures and coding rubrics, laying the groundwork for human coding. A team of experienced teachers and instructional coaches then manually codes 1,000 selected lesson plans, evaluating them against the established criteria for curriculum alignment and pedagogical efficacy. This step provides a nuanced understanding of what constitutes quality in K-12 mathematical instruction, ensuring the machine learning (ML) algorithms, trained on these evaluations, are based on a well-defined framework and expert-endorsed training dataset. 

The research team combines machine learning algorithms with human expertise to assess lesson plans in a comprehensive study to enhance the quality of educational resources. These algorithms are specifically developed to identify essential quality markers in lesson plans, such as high cognitive levels and the presence of deep, thought-provoking questions. To augment the insights from human coding, the research team develops ML algorithms to identify patterns and markers of quality in lesson plans, offering a scalable solution for assessing the broad array of available lesson materials. Concurrently, 30 math educators are employed to rate 2,000 machine-coded lesson plans using the same rubrics developed for this study. The feedback and ratings from these teachers are instrumental in refining the algorithms.

Moreover, the research team employs multi-task classification based on two powerful pre-trained language models BERT \cite{devlin_bert_2019} and RoBERTa \cite{liu_roberta_2019}, trained on the data from these 2,000 coded lesson plans. This approach enables the algorithm to rate various aspects of lesson plan quality, providing a comprehensive assessment. By integrating indicators like cognitive complexity and question depth, the system becomes adept at recognizing high-quality educational content. Researchers typically use several evaluation methods to validate the quality ratings generated by the classification. These include cross-validation, where a subset of the data is used to test the algorithm, and comparing algorithm ratings with human experts. This rigorous validation process ensures that the machine learning assessments are both accurate and aligned with the practical standards set by experienced educators. Through this blend of AI and human judgment, the study aims to offer a robust tool for educators, guiding them toward high-quality, pedagogically sound resources in the evolving digital education landscape. Therefore, by fine-tuning AI's assessment capabilities through teachers' practical understanding of quality measures drawn from classroom experience, the study ensures that algorithms are not only theoretically sound but also practically relevant and effective. Integrating  human expertise and automation presents a powerful tool for educators, enabling them to navigate and select high-quality OERs, thereby addressing the critical need for pedagogical efficacy and curricular alignment in the digital age of education.

\subsubsection{Future direction: fine-tune LLM to generate lesson materials that are consistent with these quality measures}
The future direction of the research project involves fine-tuning LLMs to generate lesson materials that adhere to established quality measures. This initiative marks a pivotal shift from evaluating existing educational resources to actively creating new ones using cutting-edge AI technology. The primary goal of fine-tuning LLMs is to align the generated lesson materials with the high-quality benchmarks set in the study's earlier phases. Furthermore, insights from teacher evaluations of existing lesson plans will inform the iterative fine-tuning process. The continuous refinement will also involve pilot testing of the AI-generated materials in real classroom settings to gather empirical data on their effectiveness and to make context-specific adjustments. A key challenge will be bridging the gap between AI capabilities and educational expertise. Addressing this, the research emphasizes  interdisciplinary collaboration, uniting AI researchers, data scientists, and educational experts to ensure that the LLMs are trained and fine-tuned in a manner that accurately reflects the pedagogical principles and practices. The team will also continuously explore customization and adaptability of LLMs, aiming to create tailored lesson materials that cater to diverse learning styles, student needs, and curriculum requirements. Ethical considerations, particularly regarding the use of data and the potential impacts on educational equity, will remain a priority. Moreover, practical aspects like the scalability of the AI-generated materials and their integration into educational infrastructures will be key focal points. The proposed future direction with LLMs represents a forward-looking move to educational content creation, leveraging the latest advancements in AI to produce high-quality, adaptable, and effective educational resources. 

To conduct supervised learning qualitative and quantitative methodologies to offer a nuanced and comprehensive assessment of large-scale lesson plan datasets. By orchestrating a symphony of cutting-edge natural language processing (NLP) techniques with established human coding methodologies, the project will delve deep into the world of online lesson plans, specifically focusing on those readily available under Creative Commons licenses and tailored to the specific needs of middle grades mathematics education. This ambitious undertaking is driven by three core research aims: forging a conceptual framework, validating measures of quality, and illuminating the teacher's lens.

A cornerstone of the study is the collaborative construction of a robust conceptual framework defining the key dimensions of high-quality middle-grades math lesson plans. This framework will be meticulously crafted by convening an expert panel of leading mathematics education researchers and seasoned practitioners. This collaborative effort will foster a shared understanding of what constitutes effective instruction in this crucial developmental stage, laying the groundwork for subsequent evaluation endeavors.

The study will then move towards developing and rigorously validating reliable measures of these key quality dimensions. This complex process will leverage both the analytical prowess of state-of-the-art computational tools, including machine learning algorithms, and the meticulous insights gleaned from human coding practices. This dual approach will ensure the accuracy and comprehensiveness of the assessment process, enabling the project to effectively evaluate a vast collection of online lesson plans.

To gain a deeper understanding of the human element within this equation, the study will embark on an exploratory mixed-methods investigation. This research phase will shed light on the cognitive processes employed by teachers as they navigate, interpret, and ultimately select lesson plans for their students. By unveiling the critical link between lesson plan quality and its tangible impact on student learning outcomes, this innovative approach will provide invaluable insights for both researchers and practitioners alike.

The transformative potential of this project for the field of mathematics education research is undeniable. By harnessing the combined power of advanced technology and human expertise, the study will move beyond traditional assessment methods and delve into the multifaceted nature of lesson plan quality. This comprehensive framework will empower educators to confidently navigate the labyrinth of online materials, armed with reliable data-driven insights to inform their decisions for optimal student learning. In doing so, the project paves the way for a future where evidence-based practices reign supreme, propelling mathematics education to new heights of effectiveness and student success.

\section{Conclusion and Future Directions}
This paper summarizes this emerging field of adopting AI/ML powered textual analysis for instructional improvement. We utilize the instructional core framework, to  guide the review  and discussion of existing studies pertaining to three components of teacher, student, and content, providing one to two case studies to illustrate the scenarios under which AI/ML can be implemented. AI, particularly Large Language Models and Generative AI, holds the potential to expedite the translation of research into EdTech products and redefine research methodologies. We highlight the transformative capacity of these technologies in enhancing instructional efficiency and effectiveness. By weaving together learning and instructional theories and latest technologies, we envision that AI-powered textual analysis and generation will make a significant contribution to K-12 research and practices. This paper is written to inspire dialogue, foster innovation, and chart a path toward an era of unparalleled educational excellence. Specifically, we envision the following trends and applications.

\paragraph{\textbf{Trend 1. Mutually informing relationship between AI/ML methodologies and educational domain knowledge.}} The mutually reinforcing relationship between AI/ML methodologies and the field of education is vital for their deep integration. This relationship is characterized by a bidirectional flow of insights. AI/ML advancements introduce revolutionary shifts in various educational aspects, such as reshaping teaching and learning experiences and enhancing educational tools. Conversely, educational domain knowledge is essential to ensure the relevance, usefulness, and effectiveness of AI/ML models in educational contexts. For instance, a significant research direction in AI for education is its adaptive role in supporting a diverse student body. AI can tailor personalized learning paths to align with individual students' unique needs, preferences, learning histories, and styles. By analyzing student performance, AI systems can swiftly identify areas of struggle and strength, enabling curriculum adjustments. This individualized approach enhances education accessibility and efficacy~\cite{mansouri2023full}. Importantly, human expertise in curriculum design is crucial for the success of AI applications. For effective learning trajectory recommendations, machine learning models depend on domain knowledge-derived knowledge graphs for accurate predictions~\cite{shaik2022review}. Additionally, AI tools help reduce teachers' administrative burdens, freeing them to focus more on instruction. For example, various higher education institutions have adopted commercial AI-based software like Gradescope and Canvas for grading and attendance monitoring. However, adopting these tools without proper human oversight and modification can introduce unnoticed biases and potentially harm student-teacher rapport. This challenge requires researchers and practitioners aiming to advance AI in education to revise the field's ontology through novel methodologies and establish a robust feedback loop between educational domain knowledge and AI/ML models. The effectiveness of such feedback loops is essential for continuous improvement and refinement during development and implementation.

Furthermore, as AI integration in education progresses, addressing ethical considerations becomes increasingly important. Incorporating human elements and domain knowledge into AI/ML algorithms is a promising approach to ensure fairness, accountability, transparency, and explainability. Additionally, involving educational stakeholders in the research and development process is essential for ensuring the inclusivity and cultural responsiveness of AI-driven tools and their applications. An effective AI/ML tool should aim to bridge educational gaps and tackle some of the most pressing challenges in education. The synergy between AI's proficiency in pattern recognition and automation and human insight in context-specific understanding has the potential to foster an adaptive, inclusive, and equitable educational ecosystem, which can empower all learners and educators by providing high-quality, personalized educational resources without bias across socio-economic backgrounds, highlighting the technology's role in promoting educational equity.

\paragraph{\textbf{Trend 2. An increasing role of multi-modal generative AI models in educational research for instructional improvement.}} Generative AI focuses on generating human-like content based on patterns and context, as well as training data. While traditional NLP primarily involves tasks such as text classification, sentiment analysis, and information retrieval, where the emphasis is on extracting specific information or insights from text. For generative AI, it is capable of completing both creative and non-creative tasks with good accuracy \cite{gozalo-brizuela_chatgpt_2023}. Due to the nature of its construction, generative AI is highly flexible in providing responses and finishing tasks. It can offer relevant and consistent answers without specific training, enabling possibilities for wide range usage and scalability. Beyond this, there are also multiple specifically trained models like Galactica that can store, combine and reason about scientific knowledge, trained on a large scientific corpus of papers, reference material, knowledge bases and many other sources \cite{taylor_galactica_2022}. However, despite its abilities to address various tasks and problems in different fields, from education to medical and management \cite{zhang_one_2023}, generative AI like ChatGPT can have limitations and failures including reasoning, factual errors, math, coding, and bias \cite{borji_categorical_2023}. Another aspect that can be both a strength and a limitation is that generative AI requires substantial computational resources, especially for large-scale models, making it difficult to implement under certain conditions. On the other hand, for traditional NLP, it is easier to tackle specific tasks, but may struggle with understanding nuanced context and generating contextually rich responses.  When focusing on specific patterns or features, traditional NLP can provide more interpretable, streamlined outputs with clearer insights into how decisions are made, making it easier to understand model behavior. Additionally, traditional NLP usually does not require intensive computational resources, making it more adaptable for condense works.

Adding on to that, there are still multiple challenges in the field that, once properly addressed, can significantly benefit this area of work. First of all, the further development of AI/ML in the field of education is still in urgent need of high-quality datasets. These datasets are large-scale, annotated datasets that are labeled based on domain knowledge of high quality teaching and learning. By feeding such data to the current AI/ML applications, NLP, and LLMs, researchers and technologists can fine-tune the existing models to make them more suitable for education-specific issues. Furthermore, if linked to school contexts, student backgrounds, and learning outcomes, AI/ML can be even more advanced toward consistently and accurately assisting the instructional performance as indicated by \citet{elmore_improving_2008}'s instructional core framework. As demonstrated in case studies, in many research projects, human experts and the process of human coding are still vital protocols that need to be completed to ensure quality responses that align with the intent of the studies. Thus, reinforcement learning and human feedback are essential to inform the process. This can also integrate teachers and educators as researchers who can actively participate in the research process, providing rigorous interactions between research and practice. Another important aspect is ethical artificial intelligence, aiming to protect privacy, increase data representativeness, address algorithm bias, establish evaluation criterion, and regulate data use. Internally, current language modeling primarily emphasizes effectiveness on standard benchmarks and efficiency, with less emphasis on reliability and practical effectiveness, and with deficiencies in constant temporal updates \cite{zhuo_red_2023}. Externally, without responsibly constructed training data and properly protected privacy data, LLMs can easily leak private information in generation \cite{carlini_extracting_2021}. During deployment, it's observed that Language Models (LLMs) may be manipulated to produce malicious content or decisions by unethical users \cite{weidinger_ethical_2021}, indicating that even internally ethical language models can be used unethically by producers and users.

\paragraph{\textbf{Trend 3. Human centered learning, with research centered on amplifying human creativity.}} Human-centered learning within the sphere of AI in education emphasizes the utilization of AI to amplify human creativity. This approach acknowledges the essential value of human experience in both the educational processes and AI-tool development, positioning AI as a catalyst for fostering creativity and developing skills. Diverting from the previous trends that focus on technology improvement as the driver of changes in the field of education, this trend emphasizes technology's role in enabling human potential through encouraging critical thinking, nuanced problem-solving, and collaboration as well as fostering imagination, curiosity, and innovative thinking. 

To realize the vision of AI in education, promoting AI literacy alongside AI integration is essential~\cite{wang2023k, velander2023artificial}. Several key areas require attention in future research. Future studies should investigate how AI can transcend the learning experience in traditional classroom settings during school hours. This involves examining AI's role in facilitating human cognition and behavioral development through adaptive algorithms and flexible accessibility. The appropriate integration of AI-based tools into current educational frameworks is required to maximize AI's capabilities in enhancing the learning process. Determining the optimal way to incorporate AI-based methodologies into existing curricula is an area that needs exploration, particularly in facilitating human critical thinking, problem-solving skills, and emotional intelligence through human-AI collaboration.

In addition to integration, developing AI literacy among students and educators is vital. AI literacy is imperative to prepare students for the demands of the future labor market. There is a need for well-designed AI curricula that equip students with the necessary skills to enter the AI-augmented labor force. These curricula should not only explain how AI functions but also address the ethical considerations of its implications and applications. Moreover, educators play a pivotal role in human-centered learning, both in delivering curricula and safeguarding students' learning experiences. As such, teachers and administrators need to be prepared for teaching in the digital age. Research on the design of professional development programs is indispensable. These programs should focus on integrating AI tools into improving teaching practices, innovating pedagogical strategies, and understanding the ethical use of AI in various educational settings. 

For future research in human-centered learning, understanding the human experience in learning about AI, as well as learning with AI, is crucial for cultivating human potential and creativity. This understanding is key to fostering a more dynamic and innovative educational landscape. 

\paragraph{\textbf{Trend 4. Building computational infrastructure, including labeled data and access to high computational capacity, which researchers can utilize for further instructional improvement.}} Large language models sometimes suffer the issue of emergent abilities, a phenomenon that unpredictably emerges. Researchers consider an ability to be emergent if it is not present in smaller models but is present in larger models \cite{wei_emergent_2022}. The existence of emergent abilities may hinder the scaling of language models which in turn restrict the development of the field. Additionally,  we are running out of the resources to train AI/ML models. It is also suggested that high-quality language data is likely exhausted before 2026, and low-quality language and image data could be exhausted by 2060 \cite{villalobos_will_2022}. Without sufficient training data, all applications related to artificial intelligence and natural language processing will be facing a severe bottleneck that restrains further advancement. Additionally, with the accelerating and significant increase of parameter magnitudes in ML models, the limitation of hardware capabilities would pose the risk of discontinuities in processing the growing model sizes \cite{villalobos_machine_2022}. To address the issue, researchers have proposed ways to reach efficient computation over LLMs, aiming to identify the optimal model size as well as number of tokens. Current models may suffer underperformance, not utilizing the computing budget to the fullest. By evaluating the balance between model sizes and training resources, LLMs can benefit from effective computational allocation \cite{hoffmann_training_2022}. 

In envisioning the future development of AI/ML methods, especially LLMs, it is crucial to establish a concrete foundation for technological capacities. This should be considered from both the hardware and software perspectives. Lying the groundwork for future development would provide us with more potential as well as more freedom to explore the application and integration of AI/ML in education.


\bibliographystyle{ACM-Reference-Format}
\bibliography{Bibliography.bib}


\end{document}